\documentclass[11pt,a4paper]{article}
\usepackage[hyperref]{acl2021}
\usepackage{times}
\usepackage{latexsym}
\usepackage{multirow}
\usepackage[colorinlistoftodos,prependcaption,textsize=tiny]{todonotes}
\usepackage[normalem]{ulem}
\usepackage{enumitem}
\useunder{\uline}{\ul}{}

\usepackage{graphicx}
\usepackage[caption=false]{subfig}

\usepackage{microtype}

\aclfinalcopy 
\def\aclpaperid{2297} 

\makeatletter
\def\blfootnote{\gdef\@thefnmark{}\@footnotetext}
\makeatother
\title{Every Bite Is an Experience: Key Point Analysis of Business Reviews}

\author{ Roy Bar-Haim, Lilach Eden, Yoav Kantor$^{*}$, Roni Friedman, Noam Slonim\\
IBM Research\\
\texttt{\{roybar,lilache,yoavka,roni.friedman-melamed,noams\}@il.ibm.com}}

\date{}

\begin{document}
\maketitle
\begin{abstract}
Previous work on review summarization focused on measuring the sentiment toward the main aspects of the reviewed product or business, or on creating a textual summary. These approaches provide only a partial view of the data: aspect-based sentiment summaries lack sufficient explanation or justification for the aspect rating, while textual summaries do not quantify the significance of each element, and are not well-suited for representing  conflicting views. Recently, \emph{Key Point Analysis (KPA)} has been proposed as a summarization framework that provides both textual and quantitative summary of the main points in the data. We adapt KPA to review data by introducing  \emph{Collective Key Point Mining} for better key point extraction; integrating sentiment analysis into KPA; identifying good key point candidates for review summaries; and leveraging the massive amount of available reviews and their metadata. We show empirically that these novel extensions of KPA substantially improve its performance. We demonstrate that promising results can be achieved without any domain-specific annotation, while human supervision can lead to further improvement. 
\end{abstract}
\section{Introduction}
\blfootnote{
    \hspace{-0.2cm}
		$^{*}$First three authors equally contributed to this work.
}
\begin{table*}[tp]
\begin{small}
    \centering
    \begin{tabular}{|l|r||l|r|}
    \hline
    \textbf{Positive Key Points} & \textbf{\% Reviews} &\textbf{ Negative Key Points} & \textbf{\% Reviews} \\	
    \hline
    \hline
    Amazingly helpful and friendly staff.	&8.6\%	& Cons: poor customer service &	9.8\% \\
    Modern furnishings and very clean.&	6.3\%&	Food is way over priced. &	3.5\% \\
    The views are incredible. &	5.2\% &	Buffet was extremely disappointing. &3.4\% \\
    The historic building is beautiful.	& 4.9\%  &	Plus it's disgusting and unsanitary. &	3.3\% \\
    Rooms are nice and comfortable. &3.8\% &	Employees are rude. &	3.2\% \\
    The rooftop pool/patio is superb.	& 3.6\% &	Rooms had a foul odor. &	3.1\% \\
    Luxurious and spacious rooms. & 2.7\%  &	Check-in took an hour. &	3.0\% \\
    The decor is very elegant. & 2.6\%  &	Staff unhelpful and uncaring. &	2.6\% \\
    The food here is excellent. & 2.4\%  & Building is very dated. & 2.3\%\\
    Great location - walkable to anything. & 2.2\%  & Our room had mechanical issues. & 1.8\%\\
    \hline
    \end{tabular}
    \caption{A sample summary produced by our system: Key Point Analysis of an hotel with 2,662 reviews from the Yelp dataset. Top 10 positive and negative key points are shown. The balanced mixture of positive and negative key points in this summary correlates with the hotel's middling rating of 3.25 stars.}
    \label{tab:kpa_example}
  \end{small}
\end{table*}
\begin{table*}[tp]
\begin{small}

\centering
    \begin{tabular}{|p{0.42\textwidth}||p{0.42\textwidth}|}
    \hline
\textbf{Key Point: The views are incredible.} & \textbf{Key Point: Cons: poor customer service} \\
\hline
    \hline

The scenery is amazing. & Service horrible   from start to finish. \\
Great view too, of the Bellagio fountains. & The front desk was   so rude to us. \\
I love this place for the scenery. & The people that   check you in suck. \\
Great room overlooking the pool. & The guy at check in   was far from friendly. \\
All were beautifully appointed and had great views of the strip. & Probably one of the worst customer experiences. \\
\hline 
\end{tabular}
\caption{
\label{tab:sample_matches} Sample matches of sentences to key points.}
  \end{small}
\end{table*}
With their ever growing prevalence, online opinions and reviews have become essential for our everyday decision making. We turn to the wisdom of the crowd before buying a new laptop, choosing a restaurant or planning our next vacation. However, this abundance is often overwhelming: reading hundreds or thousands of reviews on a certain business or product is impractical, and users typically have to rely on aggregated numeric ratings, complemented by reading a small sample of reviews, which may not be representative. The vast majority of available information is left unexploited.

Opinion summarization is a long-standing challenge, which has attracted a lot of research interest over the past two decades. Early works \cite{hu:kdd04,Gamon:2005,snyder-barzilay-2007-multiple,Blair-goldensohn:2008, titov-mcdonald-2008-joint} aimed to extract, aggregate and quantify the sentiment toward the main aspects or features of the reviewed entity (e.g.,  \emph{food}, \emph{price}, \emph{service}, and \emph{ambience} for restaurants). Such aspect-based sentiment summaries provide a high-level, quantitative view of the summarized opinions, but lack explanations and justifications for the assigned scores \citep{ganesan-etal-2010-opinosis}.

An alternative line of work casts this problem as multi-document summarization, aiming to create a textual summary from the input reviews \cite{carenini-etal-2006-multi,ganesan-etal-2010-opinosis,pmlr-v97-chu19b,brazinskas-etal-2020-unsupervised}. While such summaries provide more detail, they lack a quantitative view of the data. The salience of each element in the summary is not indicated, making it difficult to evaluate their relative significance. This is particularly important for the common case of conflicting opinions.
In order to fully capture the controversy, the summary should ideally indicate the proportion of favorable vs. unfavorable reviews for the controversial aspect. 

Recently, \emph{Key Point Analysis (KPA)} has been proposed as a novel extractive summarization framework that addresses the limitations of the above approaches \citep{Barhaim:ACL2020,Barhaim:EMNLP2020}. KPA extracts the main points discussed in a collection of texts, and matches the input sentences to these \emph{key points (KPs)}. The salience of each KP corresponds to the number of its matching sentences.  The set of key points is selected out of a set of candidates - short input sentences with high argumentative quality, so that together they achieve high coverage, while aiming to avoid redundancy. The resulting summary provides both textual and quantitative views of the data, as illustrated in  Table~\ref{tab:kpa_example}. Table~\ref{tab:sample_matches} shows a few examples of matching sentences to KPs.

Originally developed for argument summarization, KPA has also been applied to user reviews and municipal surveys, using the same supervised models that were only trained on argumentation data, and was shown to perform reasonably well. However, previous work only used KPA ``out-of-the-box'', and did not attempt to adapt it to different target domains. 

In this work we propose several improvements to KPA, in order to make it more suitable to review data, and in particular to large-scale review datasets:
\begin{enumerate}[topsep=4pt,itemsep=1pt,parsep=4pt]
\itemsep0.1em 
\item We show how the massive amount of reviews available in datasets like Amazon and Yelp, as well as their meta-data, such as numeric rating, can be leveraged for this task. 
\item We integrate sentiment classification into KPA, which is crucial for analyzing reviews.
\item We improve key point extraction by introducing \emph{Collective Key Point Mining}: extracting a large, high-quality set of key points from a large collection of businesses in a given domain. 
\item We define the desired properties of key points in the context of user reviews, and develop a classifier that detects such key points. \end{enumerate}
 
We show empirically that these novel extensions of KPA substantially improve its performance. We demonstrate that promising results can be achieved without any domain-specific annotation, while human supervision can lead to further improvement. 
Overall, this work makes a dual contribution: first, it proposes a new framework for review summarization. Second, it advances the research on KPA, by introducing novel methods that may be applied not only to user reviews, but to other use cases as well.

\section{Background: Key Point Analysis}
\label{sec:background}
KPA was initially developed for summarizing large argument collections \citep{Barhaim:ACL2020}. KPA matches the given arguments to a set of \emph{key points (KPs)}, defined as high-level arguments. The set of KPs can be either given as input, or automatically extracted from the data. The resulting summary includes the KPs, along with their salience, represented by the number (or fraction) of matching arguments.  The user can also drill down from each KP to its associated arguments. 

\citet{Barhaim:EMNLP2020} proposed the following method for automatic extraction of KPs from a set of arguments, opinions or views, which they refer to as \emph{comments}: 
\begin{enumerate}[topsep=4pt,itemsep=1pt,parsep=4pt]
    \itemsep.1em
    \item Select short, high quality sentences as \emph{KP candidates}.
    \item Map each comment to its best matching KP, if the match score exceeds some threshold $t_{match}$.
    \item Rank the candidates according to the number of their matches.
    \item Remove candidates that are too similar to a higher-ranked candidate\footnote{That is, their match score with that candidate exceeds the threshold $t_{match}$.}.
    \item Re-map the removed candidates and their matched comments to the remaining candidates.
    \item Re-sort the candidates by the number of matches and output the top-$k$ candidates.
\end{enumerate}    

Given a set of KPs and a set of comments, a summary is created by mapping each comment to its best-matching KP, if the match score exceeds $t_{match}$.

The above method relies on two models: a \emph{matching model} that assigns a match score for a \emph{(comment, KP)} pair, and a \emph{quality model}, that assigns a quality score for a given comment. The matching model was trained on the \emph{ArgKP} dataset, which contains 24K \emph{(argument, KP)} pairs labeled as matched/unmatched. The quality model was trained on the \emph{IBM-ArgQ-Rank-30kArgs} dataset, which contains quality scores for 30K arguments \cite{gretz2019largescale}\footnote{Both datasets are available from \url{https://www.research.ibm.com/haifa/dept/vst/debating_data.shtml}}. The arguments in both datasets support or contest a variety of common controversial topics (e.g., \emph{``We should abolish capital punishment''}), and were collected via crowdsourcing.

\citeauthor{Barhaim:EMNLP2020} showed that models trained on argumentation data not only perform well on arguments, but also achieve reasonable results on other domains, including survey data and sentences taken from user reviews. However, they did not attempt to adapt KPA to these domains. In the following sections we look more closely at applying KPA to business reviews.

\section{Data and Task}
In this work we apply KPA to business reviews from the Yelp Open Dataset\footnote{\url{https://www.yelp.com/dataset}}. The dataset contains about 8 million reviews for 200K businesses. Each business is classified into multiple categories. \textsc{Restaurants} is by far the most common category, comprising the majority of the reviews. Besides restaurants, the dataset contains a wide variety of other business types, from \textsc{Nail Salons} to \textsc{Dentists}. We focus on two business categories in our experiments: \textsc{Restaurants} (4.9M reviews) and \textsc{Hotels} (258K reviews). We will henceforth refer to these business categories as \emph{domains}. Each review includes, in addition to the review text, several other attributes, most relevant for our work is the ``star rating'' on a 1-5 scale. 

We filtered and split the dataset as follows. First, we removed reviews with more than 15 sentences (10\% of the reviews). Second, we removed businesses with less than 50 reviews. The remaining businesses were split into Train, Development (Dev) and Test set, as detailed in Table~\ref{tab:split}. 
\begin{table}
\begin{center}
\begin{small}
\begin{tabular}{|l|c|r|}
\hline
&\ Businesses (\%) & Reviews\\
\hline
Train & 25\% & 1,289,754 \\
Dev &25\% & 1,338,123\\
Test& 50\%  & 2,622,054\\
\hline
\end{tabular}
\caption{Yelp dataset split\label{tab:split}}
\end{small}
\end{center}
\end{table}

Our goal is to create a summary of the reviews for a given business. The summary would list the top $k$ positive and top $k$ negative KPs, and indicate   for each KP its salience in the reviews, represented by the percentage of reviews that match the KP. A review is matched to a KP if at least one of its sentences is matched to that KP. An example of such summary is given in Table~\ref{tab:kpa_example}. Table~\ref{tab:sample_matches} shows a few examples of matching sentences to KPs.

\section{Classification Models}
\label{sec:models}
Our system employs several classification models: in addition to the matching
and argument quality models discussed in Section~\ref{sec:background}, in this work we add a sentiment classification model and a KP quality model, to be discussed in the next sections. 

All four classifiers were trained by fine-tuning a RoBERTa-large model \citep{roberta-2019}. Prior to the fine-tuning of each classifier, we adapted the model to the business reviews domain, by pre-training on the Yelp dataset. We performed Masked LM pertraining \citep{devlin-etal-2019-bert, roberta-2019} on 1.5 million sentences sampled from the train set with a length filter of 20-150 characters per sentence. The following parameters were used: learning rate - 1e-5; 2 epochs. Training took two days on a single v100 GPU. 

The matching model was then obtained by fine-tuning the pre-trained model on the ArgKP dataset, with the parameters specified by \citet{Barhaim:EMNLP2020}. The quality model was fine-tuned following the procedure described by \citet{gretz2019largescale}, except for using RoBERTa-large instead of BERT-base, with learning rate of 1e-5.

\section{Incorporating Sentiment into KPA}
Previous work on KPA has ignored the issue of sentiment (or stance) altogether. When applied to argumentation data, it was assumed that the stance of the arguments is known, and KPA was performed separately for pro and con arguments. Accordingly, the ArgKP dataset only contains (argument, KP) pairs having the same stance. 

There are, however, several advantages for incorporating sentiment into KPA, in particular when analyzing reviews:
\begin{enumerate}[topsep=4pt,itemsep=1pt,parsep=4pt]
\itemsep0.1em 
\item Separating positive KPs from negative ones makes the summaries more readable.
\item Filtering neutral sentences, which are mostly irrelevant, may improve KPA quality.
\item Attempting to match only sentences and KPs with the same polarity may reduce both matching errors and run time.
\end{enumerate} 

We developed a sentence-level sentiment classifier for Yelp data by leveraging the abundance of available star ratings for short reviews. We extracted from the entire train set reviews having at most 3 sentences and 64 tokens. Reviews with 1-2, 3 and 4-5 star rating were labeled as negative (\textsc{neg}, 20\% of the reviews), neutral (\textsc{neut}, 11\%) and positive (\textsc{pos}, 69\%), respectively.  The reviews were divided into a training set, comprising 235,481 reviews, and a held-out set, comprising 26,166 reviews. 

The sentiment classifier was trained by fine-tuning the pre-trained model on the above training data, for 3 epochs. The first two rows in Table~\ref{tab:sentiment} show the classifier's performance on the held-out set.  

Since we ultimately wish to apply the classifier to individual sentences, we also annotated a small sentence-level benchmark of 158 reviews from the held-out set, which contain 952 sentences. We selected a minimal threshold $t_s$ for predicting \textsc{pos} or \textsc{neg} sentiment. If both \textsc{pos} and \textsc{neg} predictions are below this threshold, the sentence is predicted as \textsc{neut}. The threshold was selected so that the recall of both \textsc{pos} and \textsc{neg} is at least 70\%, while aiming to maximize precision\footnote{The chosen threshold was 0.79.}. Sentence-level performance on the benchmark using this threshold is shown in the last two rows of Table~\ref{tab:sentiment}. Almost all the errors involved neutral labels - confusion between positive and negative labels was very rare.

We integrate sentiment into KPA as follows. We extract positive KPs from a set of sentences classified as positive, and likewise for negative KPs. In order to further improve precision, positive (negative) sentences are only selected from positive (negative) reviews. 

When matching sentences to the extracted KPs we filter out neutral sentences and match sentences only to KPs with the same polarity. However, at this stage we do not filter by the review polarity, since we would like to allow matching positive sentences in negative reviews and vice versa, as well as positive and negative sentences in neutral reviews.  
\begin{table}
\begin{center}
\begin{small}
\begin{tabular}{|l|l|r|r|r|}
\hline
&&\textsc{pos}&\textsc{neg}&\textsc{neut}\\
\hline
\multirow{2}{*}{Reviews}&P&0.96&0.86&0.58\\
&R&0.97&0.91&0.47\\
\hline
\multirow{2}{*}{Sentences}&P&0.82&0.81&0.48\\
&R&0.88&0.70&0.47\\
\hline
\end{tabular}
\caption{Sentiment classification results on held-out data. Precision (P) and recall (R) per class are shown, for both complete reviews and individual sentences. \label{tab:sentiment}}
\end{small}
\end{center}
\end{table}
\section{Collective Key Point Mining}
\label{sec:kp-mining}
KPA is an extractive summarization method: KPs are selected from the review sentences being summarized. When generating a summary for a business with just a few dozens of reviews, the input reviews may not have enough good KP candidates - short sentences that concisely capture salient points in the reviews. This is a common problem for extractive summarization methods, where it is often difficult to find sentences that fit into the summary in their entirety. 

We propose to address this problem by mining KPs collectively for the whole domain (e.g., restaurants or hotels). The extracted set of domain KPs is then matched to the review sentences of each analyzed business. This method can extract KPs from reviews of thousands of businesses, rather than from a single business, and therefore is much more robust. It overcomes a fundamental limitation of extractive summarization - limited selection of candidate sentences, while sidestepping the complexity of sentence generation that exists in abstractive summarization. Using the same set of KPs for each business makes it easy to compare different businesses. For example, we can rank businesses by the prevalence of a certain KP of interest. 

For each domain, we sampled 12,000 positive reviews and 12,000 negative reviews from the train set, from which positive and negative KPs were extracted, respectively\footnote{To ensure diversity over the businesses, we employed a two-step sampling process: first sampled a business and then sampled a review for the business.}. We extracted positive and negative sentences from the reviews using the sentiment classifier, as described in the previous section. We filtered sentences with less than 3 tokens or more than 36 tokens (not including punctuation), as well as sentences with less than 10 characters. The number of positive and negative sentences obtained for each domain is detailed in Table~\ref{tab:sentences}.
\begin{table}
\begin{center}
\begin{small}
\begin{tabular}{|l|r|r|}
\hline
&\multicolumn{2}{|c|}{Sentences}\\
&\textsc{pos}&\textsc{neg}\\
\hline
Restaurants&49,685 & 48,751 \\
Hotels     &49,655 & 59,552 \\
\hline
\end{tabular}
\caption{Number of positive and negative sentences extracted for KP mining in each domain. \label{tab:sentences}}
\end{small}
\end{center}
\end{table} 
We ran the KP extraction algorithm described in Section~\ref{sec:background} separately for the positive and negative sentences in each domain. We used a matching threshold $t_{match}=0.99$. The length of KP candidates was constrained to 3-5 tokens, and their minimal quality score was $t_{quality}$=0.42\footnote{The threshold was selected by inspecting a sample of the training data.}. For each run, we selected the resulting top 70 candidates.

The number of RoBERTa predictions required by the algorithm is $O($\#KP-candidates $\times$ \#sentences$)$. While the input size in previous work was up to a few thousands of sentences, here we deal with 50K-60K sentences per run. In order to maintain reasonable run time, we had to constrain both the number of sentences and the number of KP candidates. We selected the top 25\% sentences with the highest quality score. The maximal number of KP candidates was $1.5 \times \sqrt{N_s}$, where $N_s$ is the number of input sentences, and the  highest-quality candidates were selected. Each run took 3.5-4.5 hours using 10 v100 GPUs.

\section{Improving Key Point Quality}
\label{sec:kpft}
Previous work did not attempt to explicitly define the desired properties KPs should have, or to develop a model that identifies good KP candidates. Instead, KP candidates were selected based on their length and argument quality, using the quality model of \citet{gretz2019largescale}. This quality model, however, is not ideally suited for selecting KP candidates for review summarization: first, it is trained on crowd-contributed arguments, rather than on sentences extracted from user reviews. Second, quality is determined based on whether the argument should be selected for a speech supporting or contesting a controversial topic, which is quite different from our use case.

We fill this gap by defining the following requirements from a KP in review summarization:
\begin{enumerate}[topsep=4pt,itemsep=1pt,parsep=4pt]
\itemsep0.1em
\item \textsc{Validity:} the KP should be a valid, understandable sentence. This would filter out sentences such as \emph{``It's rare these days to find that!''}.
\item \textsc{Sentiment:} it should have a clear sentiment (either positive or negative). This would exclude sentences like \emph{``I came for a company event''}.
\item \textsc{Informativeness:} it should discuss some aspect of the reviewed business. Statements such as \emph{``Love this place''} or \emph{``We were very disappointed''}, which merely express an overall sentiment should be discarded, as this information is already conveyed in the star rating. The KP should also be general enough to be relevant for other businesses in the domain. A common example of sentences that are too specific is mentioning the business name or a person's name (\emph{``Byron at the front desk is the best!''}).
\item \textsc{Single Aspect:} it should not discuss multiple aspects (e.g., \emph{``Decent price, respectable portions, good flavor''}).
\end{enumerate}

As we show in Section~\ref{sec:experiments}, the method presented in the previous sections extracts many KPs that do not meet the above criteria. In order to improve this situation, we developed a new KP quality classifier. 

We created a labeled dataset for this task, as follows. We sampled from the restaurant and hotel reviews in the train set 2,000 sentences comprising 3-8 tokens and minimal argument quality of $t_{quality}$. each sentence was annotated for each of the above criteria\footnote{The guidelines are included in the appendix.} by 10 crowd annotators, using the Appen platform\footnote{https://appen.com/}. We took several measures to ensure annotation quality, following  \citet{gretz2019largescale} and \citet{Barhaim:EMNLP2020}. First, the  annotation was performed by trusted annotators, who performed well on previous tasks. Second, we employed the Annotator-$\kappa$ score \citep{toledo-etal-2019-automatic}, which measures inter annotator agreement, and removed annotators whose annotator-$\kappa$ was too low. The details are provided in the appendix. For each sentence and each criterion, the fraction of positive annotations was taken to be its confidence.   

The final dataset was created by setting upper and lower thresholds on the confidence value of each of the four criteria. Sentences that matched \emph{all} the upper thresholds were considered positive. Sentences that matched \emph{any} of the lower thresholds were considered negative. The rest of the sentences were discarded. The threshold values we used are given in the appendix. Overall, the dataset contains 404 positive examples and 1,291 negative examples.

\begin{figure}
    \begin{small}
    \centering
    \includegraphics[width=6cm]{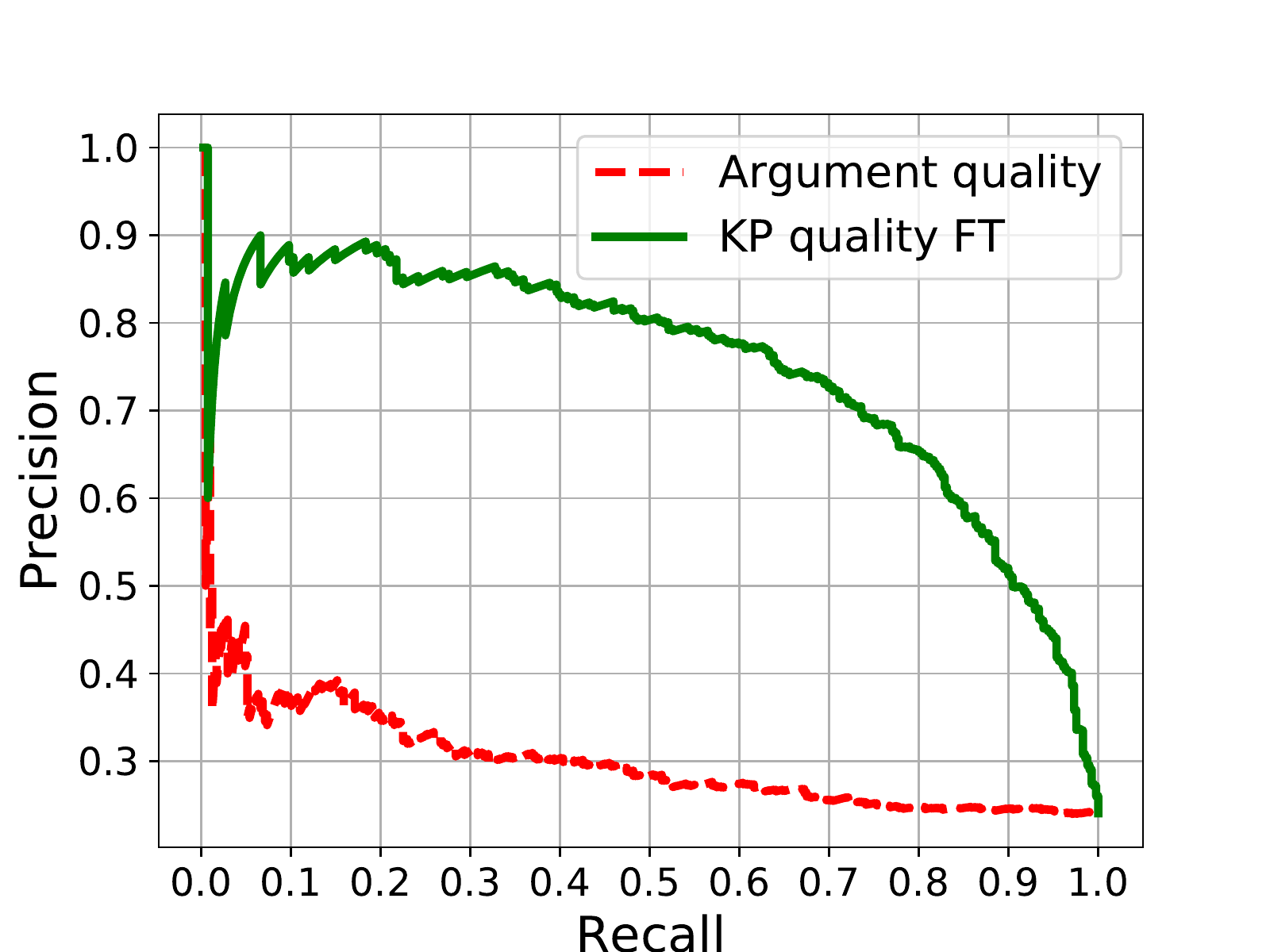}
    \caption{KP Quality Precision vs. Recall. The fine-tuned KP quality model (``\emph{KP quality FT''}) and the original argument quality model are evaluated over the KP quality labeled dataset.
    \label{fig:kpq}}
    \end{small}
\end{figure}

We trained a KP quality classifier by fine-tuning the pretrained RoBERTa model (cf. Section~\ref{sec:models}) on the above dataset (4 epochs, learning rate: 1e-05). Figure~\ref{fig:kpq} shows that this classifier (denoted \emph{KP quality FT}) performs reasonably well on the dataset, in a 4-fold cross-validation experiment.  Unsurprisingly, the argument quality classifier trained on argumentation data is shown to perform poorly on this task. 

The classifier was used to filter bad KP candidates, as part of the KP mining algorithm (Section~\ref{sec:kp-mining}). Candidates that passed this filtering were filtered and ranked by the argument quality model as before. We selected a threshold of 0.4 for the classifier, which corresponds to keeping 32\% of the candidates, with precision of 0.62 and recall of 0.82.

\section{Evaluation}
\label{sec:experiments}
\subsection{Experimental Setup}
\label{sec:setup}
Our evaluation follows \citet{Barhaim:EMNLP2020}, while making the necessary changes for our setting. Let $D$ be a domain, $K$ a set of positive and negative KPs for $D$, and $B$ a sample of businesses in $D$. Applying KPA to a business $b \in B$ using the set of KPs $K$ and a matching threshold $t_{match}$ creates a mapping from sentences in $b$'s reviews, denoted $R_b$, to KPs in $K$. By modifying $t_{match}$ we can explore the tradeoff between \emph{precision} (fraction of correct matches) and \emph{coverage}.  \citeauthor{Barhaim:EMNLP2020} performed KPA over individual sentences, and correspondingly defined coverage as the fraction of matched sentences. We are more interested in review-level coverage, since not all the sentences in the review are necessarily relevant for the summary. 

Given KPA results for $B$, $K$ and $t_{match}$, we can compute the following measures: 
\begin{enumerate}[topsep=4pt,itemsep=1pt,parsep=4pt]
    \itemsep0.1em 
    \item \emph{Review Coverage}: the fraction of reviews per business that are matched to at least one KP, macro-averaged over the businesses in $B$.
    \item \emph{Mean Matches per Review}: the average number of matched KPs per review, macro-averaged over the businesses in $B$.
\end{enumerate}

Computing precision requires a labeled sample. We create a sample S by repeating the following procedure until $N$ samples are collected:
\begin{enumerate}[topsep=4pt,itemsep=1pt,parsep=4pt]
\itemsep0.1em 
\item Sample a business $b \in B$; a review $r \in R_b$ and a sentence $s \in r$.
\item Let the KP $k \in K$ be the best match of $s$ in $K$ with match score $m$.
\item Add the tuple $[(s,k),m]$ to $S$ if $m>t_{min}$.
\end{enumerate}
The $(s,k)$ pairs in $S$ are annotated as correct/incorrect matches. We can then compute the precision for any threshold $t_{match}>t_{min}$ by considering the corresponding subset of the sample. 

We sampled for each domain 40 businesses from the test set, where each business has between 100 and 5,000 reviews. For each domain, and each evaluated set of KPs, we labeled a sample of 400 pairs.

We experimented with several configurations of KPA adapted to Yelp reviews, as described in the previous sections. These configurations are denoted by the prefix \emph{RKPA}. Each configuration only differs in the method it employs for creating the set of domain KPs ($K$):
\paragraph{\textsc{RKPA-Base}:} This configuration filters KP candidates according to their length and quality, using the quality model trained on argumentation data. In each domain, the top 30 mined KPs for each polarity were selected. 
\paragraph{\textsc{RKPA-FT}:} This configuration applies the fine-tuned KP quality model as an additional filter for KP candidates. As with the previous configuration, we take the top 30 KPs for each polarity, in each domain.
\paragraph{\textsc{RKPA-Manual}:} We also experimented with an alternative form of human supervision, where the set of automatically-extracted KPs obtained by the \textsc{RKPA-Base} configuration is manually reviewed and edited. KPs may be rephrased, redundancies are removed and bad KPs are filtered out. While this kind of task is less suitable for crowdsourcing, it can be completed fairly quickly - about an hour per domain. The task was performed by two of the authors, each working on one domain and reviewing the results for the other domain. The final set includes: 18 positive and 15 negative KPs for restaurants; 20 positive and 20 negative KPs for hotels.\footnote{The set of KPs for each configuration is provided as supplementary material.}

In addition to the above configurations, we also experimented with a ``vanilla'' KPA configuration (denoted \textsc{KPA}), which replicates the system of \citet{Barhaim:EMNLP2020}, without any of the adaptations and improvements introduced in this work. No Yelp data was used for pretraining or fine-tuning the models; key points were extracted independently for each business in the test set; and no sentiment analysis was performed. Instead of taking the top 30 KPs for each polarity, we took the top 60 KPs.

\paragraph{Sample labeling.} Similar to the KP quality dataset, the eight samples of 400 pairs (two domains $\times$ four configurations) were annotated in the Appen crowdsourcing platform. The annotation guidelines are included in the appendix. Each instance was labeled by 8 trusted annotators, and annotators with Annotator-$\kappa <0.05$ were removed (cf. Section~\ref{sec:kpft}). We set a high bar for labeling correct matches: at least 85\% of the annotators had to agree that the match is correct, otherwise it was labeled as incorrect. 

We verified the annotations consistency by sampling 250 pairs, and annotating each pair by 16 annotators. Annotations for each pair were randomly split into two sets of 8 annotations, and a binary label was derived from each set, as described above. The two sets of labels for the sample agreed on 85.2\% of the pairs, with Cohen's Kappa of 0.6\footnote{This result is comparable to \citep{Barhaim:EMNLP2020}, who reported Cohen's Kappa of 0.63 in a similar experiment.}.

\subsection{Results}
\begin{figure*}[t!]
    \centering
    
    \subfloat[Hotels]{%
        \includegraphics[width=0.5\textwidth]{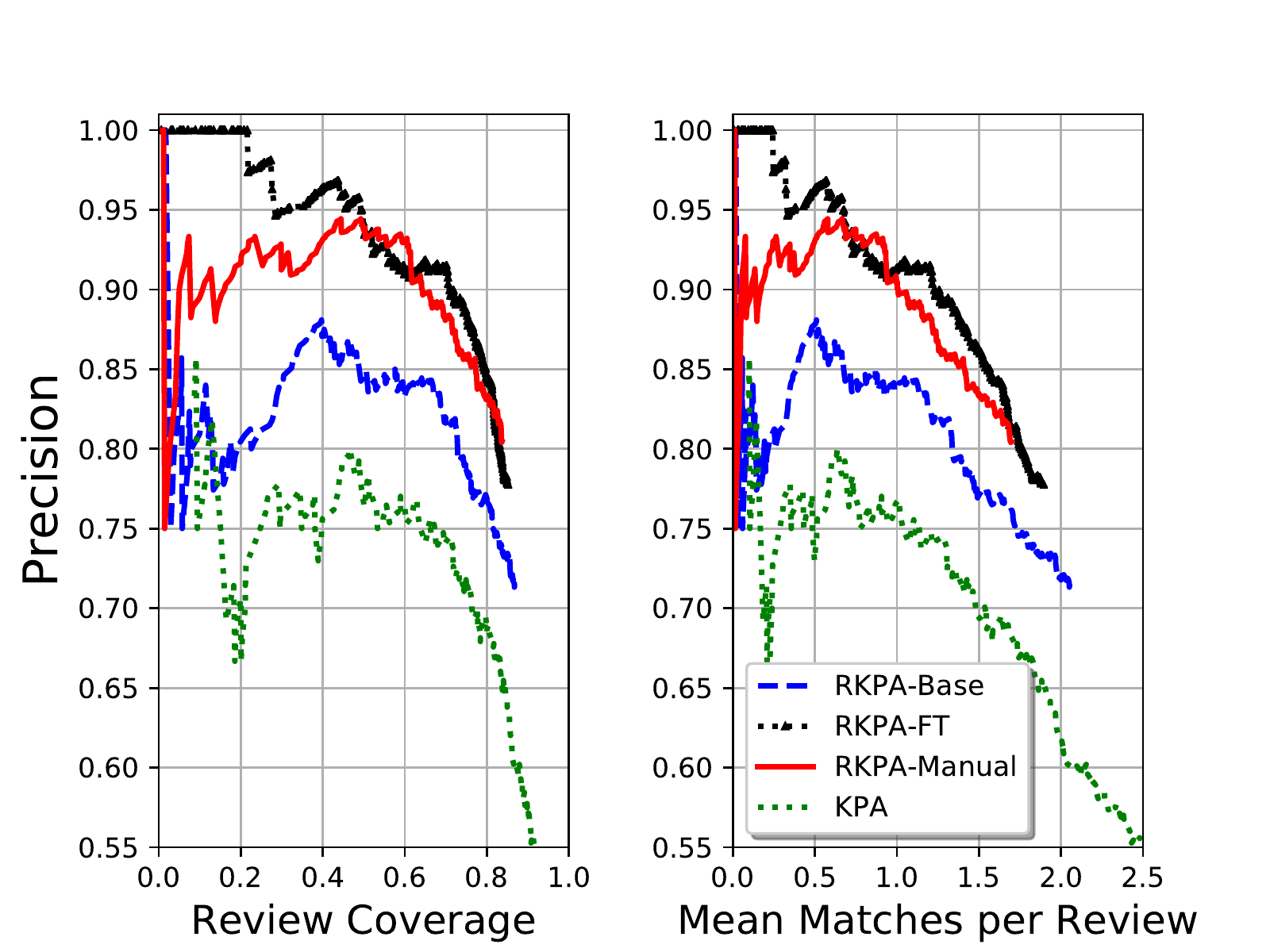}%
        \label{fig:a}%
        }%
    \hfill%
    \subfloat[Restaurants]{%
        \includegraphics[width=0.5\textwidth]{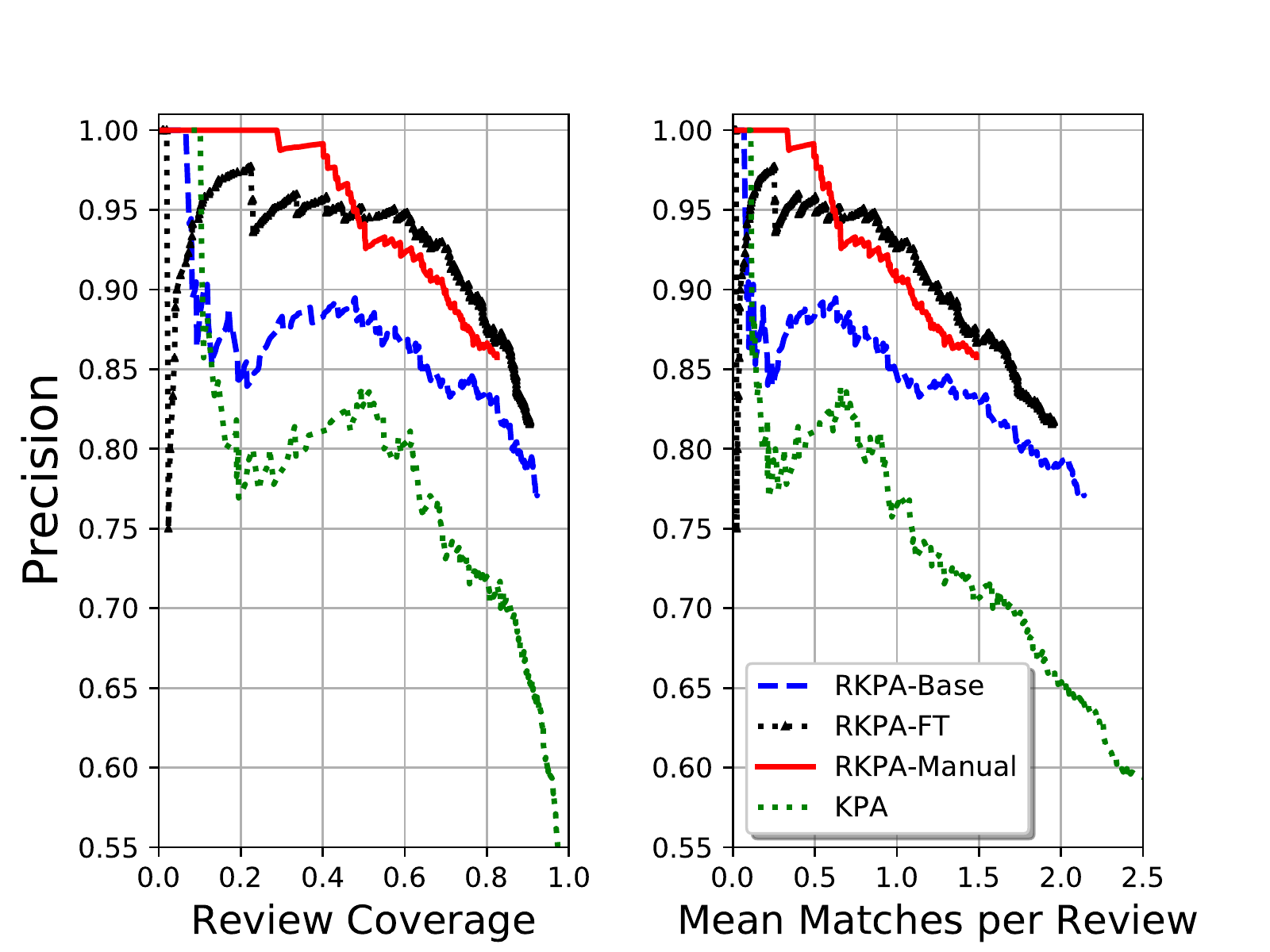}%
        \label{fig:b}%
        }%
\caption{KPA Precision vs. Coverage}
    \label{fig:precison_coverage}
\end{figure*}
Figure~\ref{fig:precison_coverage} shows the precision/coverage curves for the four configurations, where coverage is measured either as \emph{Review Coverage} (left) or as \emph{Mean Matches per Review} (right). We first note that all three configurations developed in this work outperform vanilla KPA by a large margin. 

The \textsc{RKPA-Base} configuration, which is only trained on previously-available data, already achieves reasonable performance. For example, the precision at Review Coverage of 0.8 is 0.77 for hotels and 0.83 for restaurants. Applying human supervision for improving the set of key points, either by training a KP quality model on crowd labeling (\textsc{RKPA-FT}), or by employing a human-in-the loop approach (\textsc{RKPA-Manual}) leads to substantial improvement in both domains. While both alternatives perform well, \textsc{RKPA-FT} achieves better precision at higher coverage rates.   

\begin{table*}[]
\begin{center}
\begin{small}
\begin{tabular}{|r|p{0.28\textwidth}|p{0.28\textwidth}|p{0.28\textwidth}|} 
\hline
\#&\multicolumn{1}{|c|}{\textbf{RKPA-Base}}    & \multicolumn{1}{c|}{\textbf{RKPA-FT}}         & \multicolumn{1}{c|}{\textbf{RKPA-Manual}}  \\ \hline
1&The food here is superb.                & The food here is superb.                      & Fresh and tasty ingredients          \\ \hline
2&Service and quality was excellent.      & Customer service is consistently exceptional. & Everything was delicious             \\ \hline
3&Large portions and reasonable prices.   & Service is slow and inattentive.              & Quick and polite service.            \\ \hline
4&Fantastic food, location, and ambiance. & Service was friendly and welcoming.           & Service is slow and inattentive.     \\ \hline
5&Staff is interactive and friendly.      & The food is very flavorful.                   & Staff is interactive and friendly.   \\ \hline
6&Again, flavorless and poor quality.     & Reasonably priced menu items.                 & Very affordable prices               \\ \hline
7&Ingredients where fresh and tasty.      & The restaurant is beautifully decorated.      & Atmosphere is fun and casual.        \\ \hline
8&We'll certainly be back again.          & Everything was cooked to perfection.          & The dishes are extremely overpriced. \\ \hline
9&Kevin, was rude and condescending.      & The overall ambience was pleasing.            & A lot of variety                     \\ \hline
10&Atmosphere is fun and casual.           & Staff are super nice \& attentive.            & The food was flavorless              \\ \hline
\end{tabular}
    \caption{Top 10 key points for each configuration in the restaurants domain, ranked by their number of matches in the sample. The matching threshold for each configuration corresponds to Review Coverage of 0.75.}
    \label{tab:kpa_final}
\end{small}
\end{center}
\end{table*}

Table~\ref{tab:kpa_final} shows, for each configuration in the restaurants domain, the top 10 KPs ranked by their number of matches in the sample. The matching threshold for each configuration corresponds to Review Coverage of 0.75.  For the \textsc{RKPA-Base} configuration, we can see examples of KPs that discuss multiple aspects (rows 3, 4), are too general (row 8) or too specific (row 9). These issues are much improved by applying the KP quality classifier, as illustrated by the top 10 KPs for the \textsc{RKPA-FT} configuration. 

\begin{table}[]
\begin{center}
\begin{small}
\begin{tabular}{|l|c|c|}
\hline
\multicolumn{1}{|c|}{\textbf{}} & \multicolumn{1}{c|}{\textbf{RKPA-Base}} & \multicolumn{1}{c|}{\textbf{RKPA-FT}} \\ \hline
Hotels  & 0.70        & 0.85         \\ \hline
Restaurants  & 0.62 & 0.95           \\ \hline
\end{tabular}
\caption{Key point quality assessment. For each domain and configuration, the table shows the fraction of KPs that conform to our guidelines.}
\label{tab:kpa_precision}
\end{small}
\end{center}
\end{table}
Table~\ref{tab:kpa_precision} provides a more systematic comparison of the KP quality in both configurations, based on the top 30 KPs for each polarity in each domain (120 in total per configuration). For each domain and configuration, the table shows the fraction of KPs that conform to our guidelines (Section~\ref{sec:kpft}). In both domains, KP quality is much improved for the \textsc{RKPA-FT} configuration. 

\paragraph{Error Analysis:} By analyzing the top matching errors of both domains, we found several systematic patterns of errors. The most common type of error consisted of a KP and a sentence making the same claim towards different targets, e.g. \emph{``We had to refill our own wine and ask for refills of soda.''} was matched to \emph{``Coffee was never even refilled.''}. This usually stemmed from a too specific KP and was more common in the restaurants domain. 

In some cases, a sentence was matched to an unrelated KP with a shared concept or term. For example, \emph{``Cheap, easy, and filling''} was matched to \emph{``Ordering is quick and easy''}. Polarity errors were rare but present, e.g. \emph{``However she wasn't the friendliest when she came to help us''} and \emph{``The waitress was friendly though.''}.

\section{Related Work}
Previous work on review summarization was dominated by two paradigms: aspect-based sentiment summarization and multi-document opinion summarization. 

\paragraph{Aspect-based sentiment summarization.} This line of work aims to create structured summaries that assign an aggregated sentiment score or rating to the main aspects of the reviewed entity \citep{hu:kdd04,Gamon:2005,snyder-barzilay-2007-multiple,Blair-goldensohn:2008, titov-mcdonald-2008-joint}. Aspects typically comprise 1-2 words (e.g., \emph{service, picture quality}), and are either predefined or extracted automatically. A core sub-task in this approach is \emph{Aspect-Based Sentiment Analysis:} identification of aspect mentions in the text, which may be further classified into high-level aspect categories, and classification of the sentiment towards these mentions. Recent examples are \citep{ma-etal-2019-exploring, Miao_2020, karimi2020adversarial}. 

The main shortcoming of such summaries is the lack of detail, which makes it  difficult for a user to understand why an aspect received a particular rating \citep{ganesan-etal-2010-opinosis}. Although some of these summaries include for each aspect a few supporting text snippets as ``evidence'', these examples may be considered anecdotal rather than representative. 

\paragraph{Multi-document opinion summarization.} This approach aims to create a fluent textual summary from the input reviews. A major challenge here is the limited amount of human-written summaries available for training. Recently, several abstractive neural summarization methods have shown promising results. These models require no summaries for training \citep{pmlr-v97-chu19b,brazinskas-etal-2020-unsupervised,suhara-etal-2020-opiniondigest}, or only a handful of them \citep{brazinskas-etal-2020-shot}. As discussed in the previous section, textual summaries provide more detail than aspect-based sentiment summaries, but lack a quantitative dimension. In addition, the assessment of such summaries is known to be difficult. As demonstrated in this work, KPA can be evaluated using straightforward measures such as precision and coverage.

\section{Conclusion}
We introduced a novel paradigm for summarizing reviews, based on KPA. KPA addresses the limitations of previous approaches by generating summaries that combine both textual and quantitative views of the data. We presented several extensions to KPA, which make it more suitable for large-scale review summarization: collective key point mining for better key point extraction; integrating sentiment analysis into KPA; identifying good key point candidates for review summaries; and leveraging the massive amount of available reviews and their metadata. 

We achieved promising results over the Yelp dataset without requiring any domain-specific annotations. We also showed that performance can be substantially improved with human supervision. While we focused on user reviews, the methods introduced in this work may improve KPA performance in other domains as well.

In future work we would like to generate richer summaries by combining domain level key points with ``local'' key points, individually extracted per business. It would also be interesting to adapt current methods for unsupervised abstractive summarization to generate key points.
\section*{Ethical Considerations}
\begin{itemize}
    \item Our use of the Yelp dataset has been reviewed and approved by both the data acquisition authority in our organization and the Yelp team.
    \item We do not store or use any user information from the Yelp dataset.
    \item We ensured fair compensation for crowd annotators as follows: we set a fair hourly rate according to our organization's standards, and derived the payment per task from the hourly rate by estimating the expected time per task based on our own experience. 
    \item Regarding the potential use of the proposed method - one of the advantages of KPA is that it is  transparent, verifiable and explainable - the user can drill down from each key point to it matched sentences, which provide justification and supporting evidence for its inclusion in the summary.  
\end{itemize}
\bibliographystyle{acl_natbib}
\bibliography{kp_acl21}

\begin{thebibliography}{20}
\expandafter\ifx\csname natexlab\endcsname\relax\def\natexlab#1{#1}\fi

\bibitem[{Bar-Haim et~al.(2020{\natexlab{a}})Bar-Haim, Eden, Friedman, Kantor,
  Lahav, and Slonim}]{Barhaim:ACL2020}
Roy Bar-Haim, Lilach Eden, Roni Friedman, Yoav Kantor, Dan Lahav, and Noam
  Slonim. 2020{\natexlab{a}}.
\newblock \href {https://doi.org/10.18653/v1/2020.acl-main.371} {From arguments
  to key points: {T}owards automatic argument summarization}.
\newblock In \emph{Proceedings of the 58th Annual Meeting of the Association
  for Computational Linguistics}, pages 4029--4039, Online. Association for
  Computational Linguistics.

\bibitem[{Bar-Haim et~al.(2020{\natexlab{b}})Bar-Haim, Kantor, Eden, Friedman,
  Lahav, and Slonim}]{Barhaim:EMNLP2020}
Roy Bar-Haim, Yoav Kantor, Lilach Eden, Roni Friedman, Dan Lahav, and Noam
  Slonim. 2020{\natexlab{b}}.
\newblock \href {https://doi.org/10.18653/v1/2020.emnlp-main.3} {Quantitative
  argument summarization and beyond: Cross-domain key point analysis}.
\newblock In \emph{Proceedings of the 2020 Conference on Empirical Methods in
  Natural Language Processing (EMNLP)}, pages 39--49, Online. Association for
  Computational Linguistics.

\bibitem[{Blair-goldensohn et~al.(2008)Blair-goldensohn, Neylon, Hannan, Reis,
  Mcdonald, and Reynar}]{Blair-goldensohn:2008}
Sasha Blair-goldensohn, Tyler Neylon, Kerry Hannan, George~A. Reis, Ryan
  Mcdonald, and Jeff Reynar. 2008.
\newblock Building a sentiment summarizer for local service reviews.
\newblock In \emph{NLP in the Information Explosion Era (NLPIX)}.

\bibitem[{Bra{\v{z}}inskas et~al.(2020{\natexlab{a}})Bra{\v{z}}inskas, Lapata,
  and Titov}]{brazinskas-etal-2020-shot}
Arthur Bra{\v{z}}inskas, Mirella Lapata, and Ivan Titov. 2020{\natexlab{a}}.
\newblock \href {https://doi.org/10.18653/v1/2020.emnlp-main.337} {Few-shot
  learning for opinion summarization}.
\newblock In \emph{Proceedings of the 2020 Conference on Empirical Methods in
  Natural Language Processing (EMNLP)}, pages 4119--4135, Online. Association
  for Computational Linguistics.

\bibitem[{Bra{\v{z}}inskas et~al.(2020{\natexlab{b}})Bra{\v{z}}inskas, Lapata,
  and Titov}]{brazinskas-etal-2020-unsupervised}
Arthur Bra{\v{z}}inskas, Mirella Lapata, and Ivan Titov. 2020{\natexlab{b}}.
\newblock \href {https://doi.org/10.18653/v1/2020.acl-main.461} {Unsupervised
  opinion summarization as copycat-review generation}.
\newblock In \emph{Proceedings of the 58th Annual Meeting of the Association
  for Computational Linguistics}, pages 5151--5169, Online. Association for
  Computational Linguistics.

\bibitem[{Carenini et~al.(2006)Carenini, Ng, and
  Pauls}]{carenini-etal-2006-multi}
Giuseppe Carenini, Raymond Ng, and Adam Pauls. 2006.
\newblock \href {https://www.aclweb.org/anthology/E06-1039} {Multi-document
  summarization of evaluative text}.
\newblock In \emph{11th Conference of the {E}uropean Chapter of the Association
  for Computational Linguistics}, Trento, Italy. Association for Computational
  Linguistics.

\bibitem[{Chu and Liu(2019)}]{pmlr-v97-chu19b}
Eric Chu and Peter Liu. 2019.
\newblock \href {http://proceedings.mlr.press/v97/chu19b.html} {{M}ean{S}um: A
  neural model for unsupervised multi-document abstractive summarization}.
\newblock In \emph{Proceedings of the 36th International Conference on Machine
  Learning}, volume~97 of \emph{Proceedings of Machine Learning Research},
  pages 1223--1232. PMLR.

\bibitem[{Devlin et~al.(2019)Devlin, Chang, Lee, and
  Toutanova}]{devlin-etal-2019-bert}
Jacob Devlin, Ming-Wei Chang, Kenton Lee, and Kristina Toutanova. 2019.
\newblock \href {https://doi.org/10.18653/v1/N19-1423} {{BERT}: Pre-training of
  deep bidirectional transformers for language understanding}.
\newblock In \emph{Proceedings of the 2019 Conference of the North {A}merican
  Chapter of the Association for Computational Linguistics: Human Language
  Technologies, Volume 1 (Long and Short Papers)}, pages 4171--4186,
  Minneapolis, Minnesota. Association for Computational Linguistics.

\bibitem[{Gamon et~al.(2005)Gamon, Aue, Corston-Oliver, and
  Ringger}]{Gamon:2005}
Michael Gamon, Anthony Aue, Simon Corston-Oliver, and Eric Ringger. 2005.
\newblock Pulse: Mining customer opinions from free text.
\newblock In \emph{Advances in Intelligent Data Analysis VI}, pages 121--132,
  Berlin, Heidelberg. Springer Berlin Heidelberg.

\bibitem[{Ganesan et~al.(2010)Ganesan, Zhai, and
  Han}]{ganesan-etal-2010-opinosis}
Kavita Ganesan, ChengXiang Zhai, and Jiawei Han. 2010.
\newblock \href {https://www.aclweb.org/anthology/C10-1039} {{O}pinosis: A
  graph based approach to abstractive summarization of highly redundant
  opinions}.
\newblock In \emph{Proceedings of the 23rd International Conference on
  Computational Linguistics (Coling 2010)}, pages 340--348, Beijing, China.
  Coling 2010 Organizing Committee.

\bibitem[{Gretz et~al.(2020)Gretz, Friedman, Cohen-Karlik, Toledo, Lahav,
  Aharonov, and Slonim}]{gretz2019largescale}
Shai Gretz, Roni Friedman, Edo Cohen-Karlik, Assaf Toledo, Dan Lahav, Ranit
  Aharonov, and Noam Slonim. 2020.
\newblock A large-scale dataset for argument quality ranking: Construction and
  analysis.
\newblock In \emph{AAAI}.

\bibitem[{Hu and Liu(2004)}]{hu:kdd04}
Minqing Hu and Bing Liu. 2004.
\newblock \href {https://doi.org/10.1145/1014052.1014073} {Mining and
  summarizing customer reviews}.
\newblock In \emph{Proceedings of the Tenth ACM SIGKDD International Conference
  on Knowledge Discovery and Data Mining}, KDD '04, pages 168--177, New York,
  NY, USA. ACM.

\bibitem[{Karimi et~al.(2020)Karimi, Rossi, and Prati}]{karimi2020adversarial}
Akbar Karimi, Leonardo Rossi, and Andrea Prati. 2020.
\newblock \href {http://arxiv.org/abs/2001.11316} {Adversarial training for
  aspect-based sentiment analysis with bert}.

\bibitem[{Liu et~al.(2019)Liu, Ott, Goyal, Du, Joshi, Chen, Levy, Lewis,
  Zettlemoyer, and Stoyanov}]{roberta-2019}
Yinhan Liu, Myle Ott, Naman Goyal, Jingfei Du, Mandar Joshi, Danqi Chen, Omer
  Levy, Mike Lewis, Luke Zettlemoyer, and Veselin Stoyanov. 2019.
\newblock \href {http://arxiv.org/abs/1907.11692} {Roberta: {A} robustly
  optimized {BERT} pretraining approach}.
\newblock \emph{CoRR}, abs/1907.11692.

\bibitem[{Ma et~al.(2019)Ma, Li, Wu, Xie, and Wang}]{ma-etal-2019-exploring}
Dehong Ma, Sujian Li, Fangzhao Wu, Xing Xie, and Houfeng Wang. 2019.
\newblock \href {https://doi.org/10.18653/v1/P19-1344} {Exploring
  sequence-to-sequence learning in aspect term extraction}.
\newblock In \emph{Proceedings of the 57th Annual Meeting of the Association
  for Computational Linguistics}, pages 3538--3547, Florence, Italy.
  Association for Computational Linguistics.

\bibitem[{Miao et~al.(2020)Miao, Li, Wang, and Tan}]{Miao_2020}
Zhengjie Miao, Yuliang Li, Xiaolan Wang, and Wang-Chiew Tan. 2020.
\newblock \href {https://doi.org/10.1145/3366423.3380144} {Snippext:
  Semi-supervised opinion mining with augmented data}.
\newblock \emph{Proceedings of The Web Conference 2020}.

\bibitem[{Snyder and Barzilay(2007)}]{snyder-barzilay-2007-multiple}
Benjamin Snyder and Regina Barzilay. 2007.
\newblock \href {https://www.aclweb.org/anthology/N07-1038} {Multiple aspect
  ranking using the good grief algorithm}.
\newblock In \emph{Human Language Technologies 2007: The Conference of the
  North {A}merican Chapter of the Association for Computational Linguistics;
  Proceedings of the Main Conference}, pages 300--307, Rochester, New York.
  Association for Computational Linguistics.

\bibitem[{Suhara et~al.(2020)Suhara, Wang, Angelidis, and
  Tan}]{suhara-etal-2020-opiniondigest}
Yoshihiko Suhara, Xiaolan Wang, Stefanos Angelidis, and Wang-Chiew Tan. 2020.
\newblock \href {https://doi.org/10.18653/v1/2020.acl-main.513}
  {{O}pinion{D}igest: A simple framework for opinion summarization}.
\newblock In \emph{Proceedings of the 58th Annual Meeting of the Association
  for Computational Linguistics}, pages 5789--5798, Online. Association for
  Computational Linguistics.

\bibitem[{Titov and McDonald(2008)}]{titov-mcdonald-2008-joint}
Ivan Titov and Ryan McDonald. 2008.
\newblock \href {https://www.aclweb.org/anthology/P08-1036} {A joint model of
  text and aspect ratings for sentiment summarization}.
\newblock In \emph{Proceedings of ACL-08: HLT}, pages 308--316, Columbus, Ohio.
  Association for Computational Linguistics.

\bibitem[{Toledo et~al.(2019)Toledo, Gretz, Cohen-Karlik, Friedman, Venezian,
  Lahav, Jacovi, Aharonov, and Slonim}]{toledo-etal-2019-automatic}
Assaf Toledo, Shai Gretz, Edo Cohen-Karlik, Roni Friedman, Elad Venezian, Dan
  Lahav, Michal Jacovi, Ranit Aharonov, and Noam Slonim. 2019.
\newblock \href {https://doi.org/10.18653/v1/D19-1564} {Automatic argument
  quality assessment - new datasets and methods}.
\newblock In \emph{Proceedings of the 2019 Conference on Empirical Methods in
  Natural Language Processing and the 9th International Joint Conference on
  Natural Language Processing (EMNLP-IJCNLP)}, pages 5624--5634, Hong Kong,
  China. Association for Computational Linguistics.

\end{thebibliography}
\section*{Appendices}
\appendix

\begin{table*}[]
\begin{center}
\begin{small}
\begin{tabular}{|l|l|l|}
\hline
  & \textbf{Positive}    & \textbf{Negative} \\  \hline
\textbf{Validity}   & Confidence \textgreater  0.85  & Confidence \textless  0.8\\ \hline
\textbf{Sentiment}           & Clear sentiment with confidence \textgreater 0.6 & No sentiment or sentiment confidence \textless 0.5 \\ \hline
\textbf{Informativeness}     & Informative with confidence \textgreater 0.6      & \begin{tabular}[c]{@{}l@{}@{}}Too specific\textbackslash{}not informative;\\ or doesn't refer to an aspect with\\ confidence \textgreater 0.6\end{tabular} \\ \hline
\textbf{Multiple Aspects} & Confidence \textless{}= 0.57                  & confidence \textgreater{}= 0.85   \\ \hline
\end{tabular}
\caption{Criteria for creating the key point quality dataset from crowd annotations. Sentences that match all the positive criteria are labeled as valid key points; Sentences that match any of the negative criteria are labeled as invalid key points, and the rest are excluded. }
\label{tab:kpq-table}
\end{small}
\end{center}
\end{table*}

\section{Key Point Quality Dataset}
\subsection{Annotation Guidelines} 
Below are the annotation guidelines for the KP quality annotation task:\\

In the following you will be presented with a business category and a sentence extracted from a customer review on a certain business in that category. You will be asked to answer the following questions:
\begin{enumerate}
  \item Is this a valid, understandable sentence? (Yes / No)
  \item What is the sentiment this sentence expresses toward the reviewed business or aspect of that business? (Positive / Negative / Mixed sentiment / Neutral or  unclear)
  \item Can this sentence be used to review ASPECT(S) of another business under the same category?  (No, it is too business specific / No, it does not refer to certain aspects of the business/ No, it is not informative / Yes)
  
  Note: An aspect of a business is a single attribute of its overall service/product. In hotels, for instance it could be the cleanliness of the room. In most businesses it could be the friendliness of the staff, the price, the conveniency of location etc. 
  \item Does this sentence discuss more than one independent aspect of the business? (Yes/No)
\end{enumerate}
\subsection{Quality Control}
Annotators were excluded if their Annotator-$\kappa$ score  \citep{toledo-etal-2019-automatic}, calculated for each question, was below any of these thresholds:
\begin{itemize}
    \item Question \#3 (Informativeness): 0.3
    \item Question \#4 (Multiple Aspects): 0.1
\end{itemize}
\subsection{Final Dataset Generation}
Table \ref{tab:kpq-table} shows the criteria for the inclusion of a sentence in the KP Quality dataset. Sentences that match all the Positive criteria are considered valid key points; Sentences that match any of the Negative criteria are considered invalid key points, and the rest are excluded. The confidence of a criterion denotes the fraction of positive annotations in the case of a binary choice, or the fraction of annotations for a certain label otherwise.

\section{Key Point Matching Annotation Guidelines} 
Below are the match annotation guidelines for (sentence, KP) pairs:\\

In this task you are presented with a business domain, a sentence taken from a review of a business in that domain and a key point.

You will be asked to answer the following question: does the key point match the sentence?

A key point matches a sentence if it captures the gist of the sentence, or is directly supported by a point made in the sentence.

The options are:
\begin{itemize}
\item Yes
\item No
\item Faulty key point (not a valid sentence or unclear)
\end{itemize}

\end{document}